\documentclass{article}
\pdfoutput=1
\usepackage[utf8]{inputenc}
\usepackage{enumitem}
\usepackage{hyperref}
\usepackage{tikz}
\usetikzlibrary{er,positioning}
\usepackage{epigraph}
\usepackage{forest}
\forestset{qtree/.style={for tree={align=center,inner sep=0pt}}}

\usepackage{pdflscape}

\newcommand{\myepigraph}[2]{\epigraph{\itshape #1}{---\,#2}}

\title{Arguments about Highly Reliable Agent Designs as a Useful Path to Artificial Intelligence Safety}
\author{Issa Rice \and David Manheim}
\usepackage{natbib}
\usepackage{graphicx}

\begin{document}

\maketitle

\begin{abstract}
Several different approaches exist for ensuring the safety of future Transformative Artificial Intelligence (TAI) or Artificial Superintelligence (ASI) systems \citep{yampolskiy2018artificial, bostrom2014}, and proponents of different approaches have made different and debated claims about the importance or usefulness of their work in the near term, and for future systems. Highly Reliable Agent Designs (HRAD) is one of the most controversial and ambitious approaches, championed by the Machine Intelligence Research Institute \citep{soares2017agent}, among others, and various arguments have been made about whether and how it reduces risks from future AI systems.

In order to reduce confusion in the debate about AI safety, here we build on a previous discussion by \citet{rice2020} which collects and presents four central arguments which are used to justify HRAD as a path towards safety of AI systems. We have titled the arguments (1) incidental utility, (2) deconfusion, (3) precise specification, and (4) prediction. Each of these makes different, partly conflicting claims about how future AI systems can be risky. We have explained the assumptions and claims based on a review of published and informal literature, along with consultation with experts who have stated positions on the topic. Finally, we have briefly outlined arguments against each approach and against the agenda overall.
\end{abstract}

\newpage

\section{Introduction}

To set the scene for the topic of this paper---the Highly Reliable Agent Designs (HRAD) research agenda---we first provide a broad outline of work that is happening in the field of AI safety, to situate HRAD research within the field. In this paper, we use the term \emph{AI safety} as an umbrella term referring to all of the work that goes into making AI systems do what humans want and work at least safely, and we use \emph{AI risk} as the problem that AI safety is working to solve.

\begin{figure}[htb]
  % Adjust the length here to shift the figure left or right.
   \centering
  \makebox[9cm]{%
  \small \begin{forest}, baseline, qtree
     [AI safety
      [Technical work
        [Work that attempts\\ to scale to\\ superintelligence level
          [AI control,name=control
            [AI boxing],
            [Leakproofing]
          ],
          [AI alignment
            [Interpretability,name=interp],
            [HRAD],
            [Iterated\\ amplification],
            [Value learning,name=valuelearn]
          ]
        ],
        [Near-term safety,name=nearterm
          [\phantom{wwwwwwwwwii},phantom],
          [\phantom{wwwww},phantom],
          [Making self-driving\\ cars crash less]
        ]
      ],
      [Non-technical work
        [Governance],
        [AI strategy]
      ]
    ]
     \draw (control) -- (interp);
     \draw (nearterm) -- (valuelearn);
  \end{forest}%
  }%
  \caption{Mapping the approaches to AI safety}
  \label{fig:var-hier}
\end{figure}
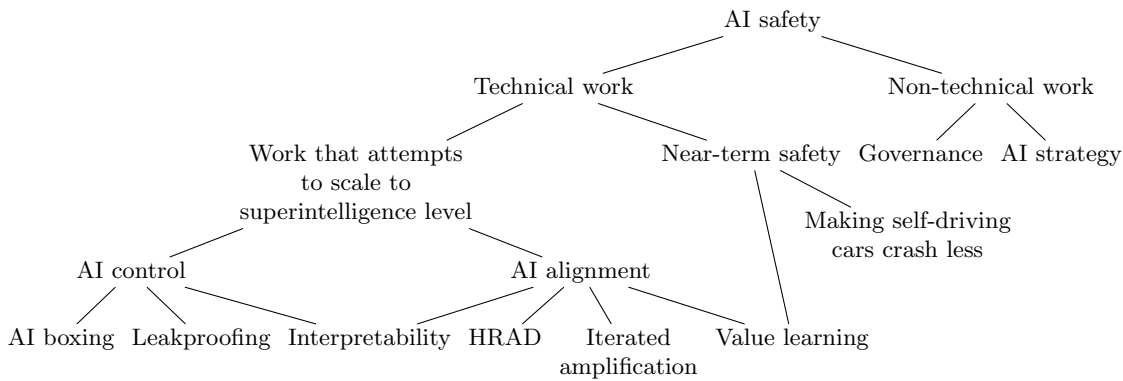

AI safety includes both technical and non-technical (e.g.\ governance\footnote{For examples of work in scalable or AI-assisted governance, see work such as \citet{evans2021truthful} and certain parts of \citet{dafoe2018ai}.}) work, as well as both work that focuses on near-term accidents and work that attempts to scale to future smarter-than-human systems including all the way to superintelligent AI. (Note that the sub-field we are discussing focuses more on risks from smarter-than-human AI compared to near-term risks.) The term AI-safety also encompasses technical approaches that attempt to ensure that humans retain the ability to control such systems or to restrict the capabilities of such systems (\emph{AI control}),\footnote{See discussion of leakproofing in \citet{chalmers2009singularity}, discussion of ``capability control methods'' in \citet{bostrom2014}, and an update and expansion of the former by \citet{yampolskiy2016leakproofing}.} as well as approaches that aim to align systems with human values---that is, the technical problem of figuring out how to design, train, inspect, and test highly capable AI systems such that they produce all and only the outcomes their creators want (\emph{AI alignment}).\footnote{What we here refer to as ``[technical] work that attempts to scale to superintelligence'' has also (especially historically) been referred to as ``AI control'' or ``the control problem'', and while this captures the concern that ``out of control'' systems are dangerous, it is potentially misleading in implying that some person or group will constrain or manipulate future smarter than human systems---a problematic outcome not necessary if one instead focuses on AI alignment.} We summarize the above distinctions in Figure \ref{fig:var-hier}.

% If we don't want the figure to run into the margins at all, we can use the "landscape" package to turn the page sideways.
%\begin{landscape}
%\end{landscape}

One of the approaches to AI alignment, the Highly Reliable Agent Designs (HRAD) agenda, is the topic of this paper. Parts of the HRAD approach to alignment has been most recently and comprehensively described in \emph{Embedded Agency}. \citep{demski2019embedded} This agenda was proposed by, and is mainly pursued by, researchers at the Machine Intelligence Research Institute (MIRI). According to one proponent, ``the HRAD agenda aims to gain a deeper, more general, and less brittle understanding of intelligence and optimization, by analyzing dilemmas where traditional formal models of agents break down. In particular, HRAD problems share a common theme of trying to generalize formal agent models to settings where the agent is embedded in its environment and must reason about itself like it reasons about other physical systems---e.g., by trying to remove traditional simplifying assumptions like `the agent knows every deductive consequence of its beliefs' and `the only ways the agent and environment can affect each other is through simple, stable, known, crisply specified input and output channels'.'' \citep{bensinger2021correspondence} The hope is that understanding ``agency'' better will help to obtain guarantees of alignment, or will otherwise assist in aligning AI systems by making us less confused (called \emph{deconfusion}, which will be explored more fully in Sections \ref{deconfusionpro} and \ref{deconfusioncon}). \citep{soares2017agent, soares2018update} The specific way in which HRAD helps with alignment has been vague, and clarifying this is the subject of the rest of the paper: each of the ``cases'' described below will give a model, either one that some proponent of HRAD research has given about how HRAD research leads to more aligned AI systems, or one that a critic of HRAD research has given when critiquing the agenda.

Various arguments for how HRAD will help to align AI systems have been discussed and critiqued privately or on online forums, but the discussions have occurred in different venues---and no single resource lists the arguments or the reasoning behind them. Not only that, but existing discourse is not clear about what the disagreements are. Partly as a result, researchers seem to disagree about the mechanism by which HRAD could plausibly reduce risks of misalignment. Until it is clearer what HRAD is, and depending on  that,  why it is useful or necessary, discussions about why to pursue or not pursue it remain vague even in extant private or forum discussions.

The goal of this paper is therefore to clearly articulate the different arguments for HRAD research which have so far been put forward, and implicitly what the agenda is required to accomplish in order for it to be successful. Below, we will attempt to explain the precise reasons given for how HRAD research will help with alignment, and disambiguate each of these arguments from others---which is especially important since they have been confused, and sound similar unless the argument is explained precisely.

\subsection{Arguments for HRAD}

The arguments or ``cases'' we will cover in this paper are: (1) incidental utility, which says that HRAD research increases our understanding of some topics relevant to AI safety, so produces some value (without trying to claim that HRAD research is the most important agenda to be pursuing); (2) deconfusion, which says that becoming less confused about agency will help to align AI systems and that without becoming less confused there is no hope of aligning AI systems; (3) precise specification (a subset of deconfusion), which says that it is possible to come up with a (more) precise theory of rationality or agency which will help to align AI systems, and that without such a theory there is no hope of aligning AI systems; and (4) prediction (also a subset of deconfusion), which says that it is possible to come up with a precise theory of rationality that allows us to analyze and predict arbitrary AI systems, which will help to align AI systems, and that without such a theory there is no hope of aligning AI systems.

These arguments are not mutually exclusive, and in fact each argument rests on a number of sometimes shared and sometimes conflicting assumptions. For this reason, the assumptions and relevant evidence are each examined to present a clearer understanding of why this agenda is being pursued by its proponents.

\subsection{Uniqueness of the research area}

Specific research topics in the HRAD agenda include embedded or realistic world models, decision theory, logical uncertainty, robust delegation, and subsystem alignment. \citep{soares2017agent, demski2019embedded} Some of these topics are active research areas in philosophy, mathematics, and computer science. This is especially the case with decision theory, which is an active research area in academic philosophy. However while the relevant parts of academic decision theory are typically more focused on comparing causal decision theory and evidential decision theory,
HRAD work is almost entirely focused on functional decision theory, which is a new kind of decision theory. \citep{yudkowsky2018functional}

More critically, the goals of the research are, if not precisely applied, at least clearly directed towards solving a set of concrete potential problems. This means that the work is somewhat interdisciplinary, and has not been embraced or fully evaluated by any of the academic disciplines it relates to. The types of work which have been done include formalisms for logical induction \citep{garrabrant2016logical}, which addresses several problems in philosophical epistemology and probability, and Löbian cooperation \citep{lavictoire2014program, critch2019parametric}, which addresses issues in computational program analysis and game theory.

\section{Background, terminology, and jargon}

In discussions considering HRAD, a number of novel or non-standard terms or distinctions have been proposed, and these will be used in the paper to better lay out the arguments made. For this reason, we introduce several key conceptual terms, specifically: (1) levels of abstraction and precise versus imprecise theory; (2) levels of indirection;  and (3) building agents from the ground up versus understanding the behavior of rational agents and predicting roughly what they will do. Because the jargon has developed over time to discuss these ideas, each distinction relates to and builds on the previous ones.

\subsection{Abstractions, and precise versus imprecise theory}

\myepigraph{Abstraction is a way to do decomposition [of a computer program] productively by changing the level of detail to be considered. When we abstract from a problem we agree to ignore certain details in an effort to convert the original problem to a simpler one.}{\citet{liskov1986abstraction}}

\myepigraph{All non-trivial abstractions, to some degree, are leaky.}{\citet{spolsky2002}}

\noindent Abstraction, as defined in the above quote by Liskov, is a key conceptual tool in a variety of domains. One key potential drawback of abstraction in computer science, highlighted by Spolsky above, is that even abstractions which work in a given context can be limited, and limiting. An abstraction is a simplification of some part of reality which enables analysis or manipulation. When the abstraction is used for prediction, it's roughly equivalent to a theory, or a model---and Spolsky's point reduces to George Box's earlier claim, ``All models are wrong, some are useful.'' But usefulness is a matter of degree. Given that, we can distinguish between a precise and imprecise, or more and less precise, theory:

\begin{itemize}
    \item A precise theory is one which can scale to two or more levels of abstraction/indirection.
    \item An imprecise theory is one which can scale to at most one level of abstraction/indirection.
\end{itemize}

More intuitively, a precise theory is more mathematical, rigorous, and exact, like those found in pure mathematics and physics, and an imprecise theory is more akin to an analogy which is locally useful, and is still predictive, but cannot be used to build higher level models.\footnote{A perhaps useful example is found in economics, where microeconomics and macroeconomics were built largely in parallel before a movement in the mid-20th century to build microeconomic foundations of macroeconomics. To the extent that it was successful, the ability to move from personal preferences and utility functions, to theory of trade and markets, to general equilibria, to aggregate questions of supply and demand, and finally to macroeconomic phenomena like inflation,  international growth,  recessions, and business cycles would be a precise theory.}

\subsection{Levels of indirection}

Distinct from levels of abstraction is a concept we call ``levels of indirection'' which is used to consider the theory of change for AI safety. Rohin Shah has called this ``multiple levels of inference'', and discussed it in terms of necessary assumptions. 
\citep{shah2020realismcomment, shah2020realismcomment2}

If a theory of change has many assumptions, each of which needs to be true, it is weaker than one which relies on fewer assumptions. \citep{MR-114-A} Similarly, for theories, each leaky abstraction is also an assumption that the abstraction does not neglect a critical factor. And for building systems, relying on less precise theories creates additional assumptions needed for the theory to apply and the system to work.

%Causal theory of change / causal distance 

\subsection{Building versus Predicting} \label{buildingvspredicting}

The distinction between building agents from the ground up versus understanding the behavior of rational agents and predicting roughly what they will do is due to \citet{demski2020realismcomment}. The distinction was only outlined briefly, so we have elaborated on this distinction. For that reason, what we describe may be a slightly different distinction than originally intended.

\begin{itemize}
    \item \emph{Building agents from the ground up} means having a theory of optimally pursuing goals, i.e.\ rationality, that is precise enough to allow us to build human-level or smarter AI systems in a satisfying way. Here, ``satisfying way'' is somewhat complicated to explain, but is some combination of traits like the human operators having a good idea of how the AI system is reasoning internally (unlike e.g.\ current mainstream machine learning models, which tend to be uninterpretable) and where someone with security mindset \citep{yudkowsky2017security} can be confident that it is aligned. Importantly, we do not require that the AI system be built using whatever methods happen to be mainstream, but instead allow it to be built using the precise theory.
    \item \emph{Understanding the behavior of rational agents and predicting roughly what they will do} means having a theory of rationality that is so precise that one can take as input an arbitrary agent and can predict roughly how it will act. This prediction ability must hold even when the agent is implemented in incomprehensible ways like blackbox machine learning models. See also Appendix \ref{clarificationprediction}.
\end{itemize}

The distinction between a theory sufficient to build aligned AGI and one sufficient to understand all AI systems is similar to the distinction between the existential quantifier ($\exists$, ``there exists'') and the universal quantifier ($\forall$, ``for all''). In the former case, we ``only'' need a model sufficient for at least one safe AI system. To make this distinction more formal, we can consider the predicate $P(x)$ standing for ``We can understand and predict how the agent or advanced AI system $x$ will act in a satisfying way.'' Then ``building agents from the ground up'' can be seen as the claim that we can invent a theory such that there exists some $x$ for which $P(x)$ is true. The more ambitious goal of ``understanding the behavior of rational agents and predicting roughly what they will do'' requires that we can invent a theory such that for all $x$, the statement $P(x)$ is true.

In other words, the former only claims we can build some advanced AI system using the precise theory that we understand well, whereas the latter says we have to deal with whatever kind of advanced AI system that ends up being developed, including using methods we do not understand well (i.e.\ where the internal reasoning of the system is not clear to us).

\section{Arguments for each case for HRAD Research}\label{argumentspro}
% "case" corresponds to "world" from the blog post

In this section, we explain each of the ``cases'' for HRAD research, that is, different arguments for doing this kind of research, and gives citations to the people making these arguments. We first present the arguments themselves, without trying to represent the debates---but arguments against each case will be given in Section \ref{argumentscon}. These argument are, in brief:

\begin{description}
    \item[Incidental Utility:] There is value to HRAD work for alignment, but it is not directly enabling aligned AI.
    \item[Deconfusion:] HRAD work is essential to ensure we actually know what problem we are solving with alignment, but is not itself going to lead to alignment.
    \item[Precise Specification:] HRAD work will allow us to find at least one aligned AGI system.
    \item[Prediction:] HRAD work will be able to tell us how to align an arbitrary AI system, and/or whether it is possible to do so.
\end{description}

Except for the first case (incidental utility), each of the cases for HRAD research are defending the claim that HRAD research is the most important kind of AI safety research to be doing, and furthermore, that without it alignment of smarter-than-human AI systems is impossible.

The proponents of HRAD in general have sometimes used more than one argument/case, or have been ambiguous about which case they are making. Given that ambiguity, there is uncertainty not just about whether the pro-HRAD versus anti-HRAD stance is correct within a single case, but also about which case represents the main reason to pursue and/or prioritize this research agenda.

\subsection{Incidental utility}
% This corresponds to the "weaker pro-HRAD" case mentioned in the original blog post.

The incidental utility case for HRAD states that HRAD research produces value for AI safety, without reference to other technical research agendas (i.e.\ without claiming that HRAD is better than other agendas,) and without claiming that not doing enough HRAD research will lead to misalignment in AI systems. As such, this case is easier to defend, but it is also less interesting because it does not prioritize among the different technical AI safety research agendas (which is critical when trying to allocate resources to the different research agendas).

As an example of someone who has made the incidental utility case for HRAD, Wei Dai has argued that decision theory research is useful for ``[b]etter understanding potential AI safety failure modes that are due to flawed decision procedures implemented in or by AI'' \citep{dai2019purposesdecision} without trying to compare this kind of research to other kinds of research. \citep{dai2019purposesdecisioncomment}

\subsection{Deconfusion} \label{deconfusionpro}

In the context of AI safety research, the term \emph{deconfusion} was introduced at the Machine Intelligence Research Institute. Throughout the history of math and science, people went from being totally confused on topics like infinity, electromagnetism, and how humans and animals came to exist, to coming up with a theory about how to reason coherently about these concepts and answer these questions. The claim the deconfusion case makes builds on this history, and suggests that this same process will happen with concepts that currently confuse us about how the mind, agents, counterfactual reasoning, logical uncertainty, and other concepts relevant to AI research work. Furthermore, they claim that it is possible to make progress on becoming less confused by focusing on this directly. Once we become less confused about minds and agency, we will have some hope of aligning advanced AI systems, because being less confused will help to avoid, detect, and fix safety issues; help to predict or explain safety issues; help to conceptually clarify the AI alignment problem; and help us be satisfied that any AI system we build is doing what we want. In fact, proponents analogize, not having such an understanding would be like trying to build a rocket to go to the Moon without having invented calculus, much less orbital mechanics and Newtonian physics. \citep{yudkowsky2018rocket}

For this case, it is not necessary that the final product of HRAD research be a precise theory. Even if the final theory is imprecise or there is no final theory, if we are merely much less confused than we are now, that may still be good enough to help us develop methods to ensure AI systems are aligned.

A number of researchers and other staff at MIRI have made arguments quite like the deconfusion case, including
\citet{yudkowsky2016prereview},
\citet{bensinger2018comment},
\citet{demski2018realismcomment}, and
\citet{soares2017hradcomment, soares2018update}.
We give quotes in Appendix \ref{deconfusionquotes}.

\subsection{Precise specification}

The precise specification case states that the goal of HRAD research is to produce a theory of agency/rationality that is so precise that it allows us to build an agent from the ground up, i.e.\ to precisely specify how an agent will work.

For this case, deconfusion is still important (or in other words, we expect that being able to build an agent from the ground up will lead to becoming far less confused about agency), so this case is a special case of the deconfusion case. However, for this case we seek not just any kind of deconfusion, but specifically deconfusion which is accompanied by a precise theory of rationality.

As with the deconfusion case, once we become less confused about minds and agency, we will have some hope of aligning advanced AI systems. This is because being less confused will help to avoid, detect, and fix safety issues; help to predict or explain safety issues; help to conceptually clarify the AI alignment problem; and help us be satisfied that any AI system we build is doing what we want.

At least one researcher at MIRI, namely Abram Demski, has argued that precise specification is possible (without necessarily claiming that he is working toward this goal). He has stated in part that he expects that ``at some point we will be able to build agents from the ground up''. \citep{demski2020realismcomment} Demski also refers to a quote from another researcher at MIRI, Eliezer Yudkowsky, who once wrote ``Such understanding as I have of rationality, I acquired in the course of wrestling with the challenge of artificial general intelligence (an endeavor which, to actually succeed, would require sufficient mastery of rationality to build a complete working rationalist out of toothpicks and rubber bands).'' \citep{yudkowsky2006martial} Yudkowsky's quote is talking about AI research in general (rather than specifically AI safety), so it is not the same thing as the precise specification case, but it is still pointing to the same kind of output that the precise specification case is seeking.

Among the critics of HRAD research within AI safety, some of them seem to be critiquing the precise specification case. \citep{shah2020realismcomment, shah2020realismcomment2} We will cover these critiques in Section \ref{argumentscon}.

\subsection{Prediction}

The prediction case states that the goal of HRAD research is to produce a theory of agency/rationality that is so precise that it directly allows us to take arbitrary agents and predict how they will work, and that having such a theory is important to understanding AI systems and solving the alignment problem.

Thus, like with the precise specification case, the prediction case is a special case of the deconfusion case. However unlike the precise specification case, here we require that the precise theory be able to predict arbitrary agents, which seems like a much more demanding requirement (as explained in Section \ref{buildingvspredicting}, precise specification is like existential quantification and prediction is like universal quantification). Also, rather than using the precise theory to become less confused, and then using that insight to help with alignment (as with the precise specification case), with the prediction case we directly make use of the precise theory to help us align AI systems. This distinction between precise specification and prediction was first made by Abram Demski. \citep{demski2020realismcomment} See also Appendix \ref{clarificationprediction} for further clarification of what prediction means.

As with the deconfusion and precise specification cases, once we become less confused about minds and agency, we will have some hope of aligning advanced AI systems. This is because being less confused will help to avoid, detect, and fix safety issues; help to predict or explain safety issues; help to conceptually clarify the AI alignment problem; and help us be satisfied that any AI system we build is doing what we want.

Although there seem to be no MIRI researchers who have made the prediction case for HRAD research, some critics of HRAD research seem to be ascribing this view to proponents of HRAD research, e.g.\ \citet{dewey2017hrad}; we will discuss these critiques in Section \ref{argumentscon}. In addition, at least one AI safety researcher, namely John Wentworth, has explicitly stated that he is working toward this prediction goal, and believes that the prediction case for HRAD is ``both sufficient and true'' (though not necessary) to defend HRAD research as the highest priority for AI alignment research. \citep{wentworth2020plausible}

\section{Arguments against HRAD} \label{argumentscon}

In this section, we go over each of the cases for HRAD research and discuss the criticisms that have been made for the case. We also briefly cover some general skepticism of AI alignment, which is relevant to all of the cases for HRAD research as well as other approaches. We describe each of the arguments in some detail, but do not  try to resolve the debate.

\subsection{Skepticism about AI alignment in general}

All of the cases for HRAD are unconvincing if the AI alignment ``problem'' in general is not a problem: why work on trying to make AI systems safe if we don't expect misaligned AI to cause trouble? We might expect working on AI safety to be useless or harmful for a long list of reasons, but here we consider two common objections: human-level or superhuman-level AI being impossible, and alignment being easy enough to happen by default.\footnote{A wider range of objections are encapsulated by these two arguments than one might at first suppose. For instance, an argument like ``the AI won't have a body so it can't hurt us'' is a claim that not connecting the AI to a robotic body suffices as a solution to AI safety, so it is a special case of AI safety being easy.} \citep{yampolskiy2021ai} The arguments for why the risk of superhuman AI is plausible or even inevitable, have been discussed in the literature \citep{bostrom2014, amodei2016concrete}, and updated discussion of the arguments for the importance of such work has been discussed more recently by \citet{ngo2019disentangling}, and we will not review them here. Some objections to those arguments exist, and we will mention three.

First, if human-level or greater AI is impossible, the alignment problem will presumably be much easier (as we only need to align weak systems like existing AI systems) or non-existent (because AI systems would not be capable of causing catastrophe, so we would only be dealing with more minor accidents).

Second, if AI is ``aligned by default'' (i.e.\ automatically aligned or very easy to align so that it happens by default), then clearly no separate investment into safety is necessary.

Finally, one prominent critic of AI existential risk who has argued along the above lines is the psychologist and linguist Steven Pinker. Pinker's reasoning is that if humans are competent enough to be able to design advanced AI systems, then they would also be competent enough to have tested it; and that any AI which is intelligent enough to cause catastrophe would also be intelligent enough to understand nuances in the goals it was instructed to perform. \citep{pinker2018enlightment, pinker2018robot} 

We will not attempt to recapitulate the general arguments further. The remaining objections to HRAD work in Section \ref{argumentscon} take the alignment problem as given, so the rest of the section will be conditional on AI safety mattering.

\subsection{Skepticism about HRAD in general, and response to incidental utility}

Now we cover objections to HRAD research more specifically. We first cover general arguments against HRAD research that don't fit in under any of the cases or that apply equally well to all or most of the cases. Since the incidental utility case is so ``weak'' (in the sense that it is easier to defend), this also means that many of these general arguments against HRAD will end up being a response to the incidental utility case as well. Thus, we won't cover the arguments against incidental utility separately in a later section.

One way to object to HRAD research is to present a different technical AI safety agenda and claim that it is better. (This won't apply to the incidental utility case since that case is not trying to compare HRAD to other research agendas, but does apply to all the other cases.) Indeed, besides HRAD there already exist several other technical agendas in AI safety. If some combination of these agendas is easier to pursue and is sufficient to ensure alignment of highly capable AI systems, then HRAD is redundant (modulo portfolio approach arguments; see the Conclusion for more about the portfolio approach). The cases for these other agendas differ based on the agenda, and explaining cases for these other agendas is outside the scope of this paper.

Another argument against HRAD in general (which also applies to the incidental utility case) is that HRAD research may be net harmful if it advances AI capabilities (leading to shorter AI timelines, meaning less time to work on alignment), or if it advances capabilities relative to alignment. Wei Dai has made this argument a few times \citep{dai2015amacomment, dai2011singularitycomment, dai2012workshopcomment3}, even giving it as a reason he personally has stopped doing decision theory research. \citep{dai2012workshopcomment, dai2012workshopcomment2}

\subsection{Response to deconfusion} \label{deconfusioncon}

There are several ways that AI safety researchers have responded to the deconfusion case.

The first is to argue that other research agendas, such as the agendas working on machine learning safety (ML safety), also involve some deconfusion, and there might be other kinds of deconfusion work too (such as work in cognitive psychology), so HRAD might not be the best path toward becoming less confused about concepts like minds, agency, and logical uncertainty. \citep{ngo2020plausible}

Another argument made against deconfusion is that there are too many levels of indirection. If the goal of HRAD is deconfusion without a precise theory as final product, it is unhelpful because  having merely an imprecise theory is insufficient for helping with alignment. \citep{shah2020realismcomment}

Finally, there are other arguments against deconfusion that apply to more restricted cases where a precise theory is sought; these will be covered in the following two sections.

\subsection{Response to precise specification}

There are two ways that the precise specification case has been dismissed. The first is to reject the existence of a precise theory of rationality. For instance, AI safety researcher Rohin Shah argues that ``MIRI's theories will always be the relatively-imprecise theories that can't scale to `2+ levels above'.'' \citep{shah2020realismcomment, shah2020realismcomment2} Paul Christiano, another AI safety researcher, seems to also make this rejection: in a post by Jessica Taylor summarizing his views, one of  the intuitions listed (intuition 18) is ``There are reasons to expect the details of reasoning well to be `messy'.'' \citep{taylor2017paulmiri} Here, ``messy'' reasoning is in contrast to the kind of theoretically elegant reasoning that a precise theory of rationality would offer, and evidently the claim is that because good reasoning is inherently messy, a precise theory cannot be found.

The second way to reject precise specification is to argue that even having a precise theory of rationality is insufficient to help align AI systems. This seems to be Open Philanthropy program officer Daniel Dewey's criticism of HRAD. For instance, he writes ``AIXI and Solomonoff induction are particularly strong examples of work that is very close to HRAD, but don't seem to have been applicable to real AI systems'' and ``It seems plausible that the kinds of axiomatic descriptions that HRAD work could produce would be too taxing to be usefully applied to any practical AI system''. \citep{dewey2017hrad} Implicit in this argument is that if AGI and/or ASI is built before HRAD is successful, the work will be irrelevant.

\subsection{Response to prediction}

The arguments against this case are essentially identical to the arguments against the previous (precise specification) case, i.e.\ one can reject the existence of a precise theory for understanding the behavior of arbitrary rational agents, or one can argue that even having such a precise theory is insufficient to help with alignment.

Since the distinction between the precise specification case and the prediction case is recent, there don't seem to be arguments in the literature specifically against one of precise specification or prediction; instead, the existing arguments are against precise theories in general. For example, \citet{dewey2017hrad} could be interpreted to be arguing against either case.

\section*{Conclusion}

% David suggested changing the boolean "and/or" statement here into a diagram: https://www.greaterwrong.com/posts/BGxTpdBGbwCWrGiCL/plausible-cases-for-hrad-work-and-locating-the-crux-in-the#Conclusion__and_moving_forward
% David: Now done, feeedback on how well it works or how to do it better is welcome.

In this paper, we explained the HRAD agenda for AI alignment. We outlined four arguments for why HRAD is either worth working on alongside other approaches, or should be a top priority for technical AI alignment research. We also covered arguments against each of these four cases, as well as arguments against AI safety and HRAD in general.

%\pagebreak

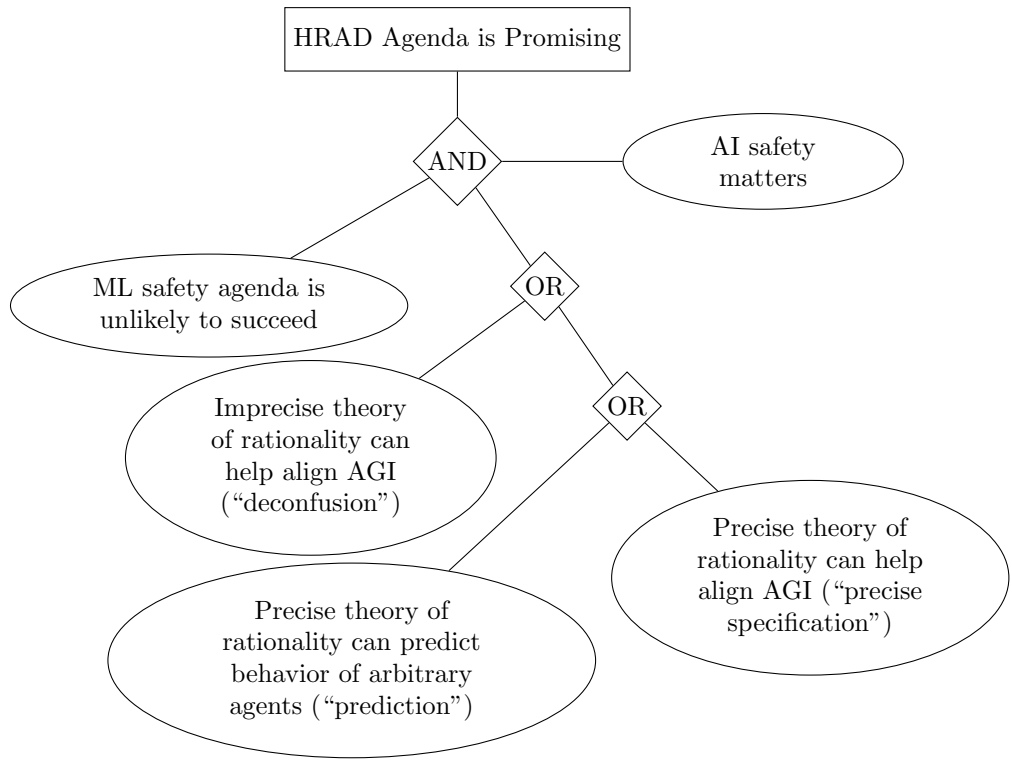
\begin{figure}[ht]
\caption[Logical Implications for Arguments Presented]{Logical Implications for Arguments Presented\footnotemark}
\vspace{3mm}
\begin{tikzpicture}[auto,node distance=1.6cm]
  \node[entity] (promising) {HRAD Agenda is Promising};
  \node[relationship] (and1) [below = of promising, yshift=10mm] {AND};
  \path (promising) edge (and1);
  \node[attribute] (safety_matters) [right= of and1, text width=2.4cm, align=center] {AI safety matters};
  \path (and1) edge (safety_matters);
  \node[attribute] (ml_unpromising) [below left = of and1, text width=3.5cm, align=center] {ML safety agenda is unlikely to succeed};
  \path (and1) edge (ml_unpromising);
  \node[relationship] (or1) [below right = of and1, xshift=-5mm] {OR};
  \path (and1) edge (or1);
  \node[attribute] (imprecise) [below left = of or1, text width=3.25cm, align=center] {Imprecise theory of \nobreak{rationality} can help align AGI (``deconfusion'')};
  \path (or1) edge (imprecise); 
  \node[relationship] (or2) [below right = of or1, xshift=-5mm] {OR};
  \path (or1) edge (or2);
  \node[attribute] (precise_align) [below right = of or2, text width=3.5cm, align=center, xshift=-8mm] {Precise theory of \nobreak{rationality} can help align AGI (``precise specification'')};
  \node[attribute] (precise_predict) [below left = of or2, text width=4.35cm, yshift=-11mm, align=center] {Precise theory of \nobreak{rationality} can predict behavior of arbitrary agents (``prediction'')};
  \path (or2) edge (precise_align);
  \path (or2) edge (precise_predict);

  % Now place a relation (ID=rel1)
  %\node[relationship] (rel1) [above right = of promising] {Relation 1};
  % Now the 2nd entity (ID=rel2)
  %\node[entity] (node2) [below right = of rel1]	{Fancy Node 2};
  % Draw an edge between rel1 and node1; rel1 and node2
  %\path (rel1) edge node {1-\(m\)} (node1)
    %edge	 node {\(n\)-\(m\)}	(node2);
\end{tikzpicture}
\end{figure}
\footnotetext{The case for HRAD can be summarized as a Boolean expression the figure illustrates:\\
(AI safety matters) and\\
(machine learning safety agenda not promising) and (\\
\phantom{W}(even an imprecise theory of rationality helps to align AGI) or\\
\phantom{W}((a precise theory of rationality can be found) and\\
\phantom{W}\phantom{W}(a precise theory of rationality can be used to help align AGI)) or\\
\phantom{W}(a precise theory to predict behavior of arbitrary agent can be found)  )\\
We don't mention the incidental utility case here as it is arguing for a different conclusion about the importance of HRAD.
}

Despite the arguments and disagreements noted, however, there seems to be agreement across the majority of people involved in these discussions that there is value in multiple approaches, and a portfolio of approaches to AI safety is widely agreed to be valuable.\footnote{See \citet{krakovna2017portfolio} for one example and elaboration.} This remains true even if we are uncertain that AI poses a risk in the future, unless we are strongly confident we can rely on the arguments that AI will be aligned by default. While the disagreements both within AI safety and between A safety proponents and skeptics are not likely to be resolved quickly, we hope that clarifying these arguments will lead to better understanding of the motives and approaches among machine learning and AI researchers generally, and more productive disagreements within the narrower field of AI safety.

\section*{Acknowledgments}

We thank Rob Bensinger, Ross Gruetzemacher, Rohin Shah, John Wentworth, and Roman Yampolskiy for feedback on a draft of this paper. Issa thanks amc, Anon User, Rob Bensinger, Vanessa Kosoy, Richard Ngo, Ben Pace, Pongo, Rohin Shah, and John Wentworth for comments on the blog post \citep{rice2020} on which this paper is based. We thank the Effective Altruism Funds Long-Term Future Fund for funding work on this paper.

\newpage

% All the URLs we have make justified text have uncomfortably large spaces, so the following will left-align just the bibliography.

\bibliographystyle{plainnat}
\bibliography{references}

\begin{thebibliography}{46}
\providecommand{\natexlab}[1]{#1}
\providecommand{\url}[1]{\texttt{#1}}
\expandafter\ifx\csname urlstyle\endcsname\relax
  \providecommand{\doi}[1]{doi: #1}\else
  \providecommand{\doi}{doi: \begingroup \urlstyle{rm}\Url}\fi

\bibitem[Amodei et~al.(2016)Amodei, Olah, Steinhardt, Christiano, Schulman, and
  Mané]{amodei2016concrete}
Dario Amodei, Chris Olah, Jacob Steinhardt, Paul Christiano, John Schulman, and
  Dan Mané.
\newblock Concrete problems in {AI} safety, 2016.
\newblock URL \url{https://arxiv.org/abs/1606.06565}.

\bibitem[Bensinger(2018)]{bensinger2018comment}
Rob Bensinger.
\newblock Comment on decision theory, 2018.
\newblock URL
  \url{https://www.alignmentforum.org/posts/uKbxi2EJ3KBNRDGpL/comment-on-decision-theory}.

\bibitem[Bensinger(2021)]{bensinger2021correspondence}
Rob Bensinger.
\newblock Private correspondence re: arguments for an hrad agenda, November
  2021.

\bibitem[Bostrom(2014)]{bostrom2014}
Nick Bostrom.
\newblock \emph{{Superintelligence: Paths, dangers, strategies}}.
\newblock Oxford University Press, Oxford, UK, 2014.
\newblock ISBN 210077252X.

\bibitem[Chalmers(2009)]{chalmers2009singularity}
David Chalmers.
\newblock The singularity: A philosophical analysis.
\newblock \emph{Science fiction and philosophy: From time travel to
  superintelligence}, pages 171--224, 2009.

\bibitem[Critch(2019)]{critch2019parametric}
Andrew Critch.
\newblock A parametric, resource-bounded generalization of {L{\"o}b’s}
  theorem, and a robust cooperation criterion for open-source game theory.
\newblock \emph{The Journal of Symbolic Logic}, 84\penalty0 (4):\penalty0
  1368--1381, 2019.

\bibitem[Dafoe(2018)]{dafoe2018ai}
Allan Dafoe.
\newblock Ai governance: a research agenda.
\newblock \emph{Governance of AI Program, Future of Humanity Institute,
  University of Oxford: Oxford, UK}, 1442:\penalty0 1443, 2018.

\bibitem[Dai(2011)]{dai2011singularitycomment}
Wei Dai.
\newblock Comment on ``some thoughts on singularity strategies'', 2011.
\newblock URL
  \url{https://www.lesswrong.com/posts/73SotZnDbsYpxfnuQ/some-thoughts-on-singularity-strategies?commentId=rXj3Q9MWofhNFh6aa}.

\bibitem[Dai(2012{\natexlab{a}})]{dai2012workshopcomment}
Wei Dai.
\newblock Comment on ``singularity summit 2011 workshop report'',
  2012{\natexlab{a}}.
\newblock URL
  \url{https://www.lesswrong.com/posts/acxdcBuZPkvnM9TNM/singularity-summit-2011-workshop-report?commentId=e8HsXCgMZcfsctk42}.

\bibitem[Dai(2012{\natexlab{b}})]{dai2012workshopcomment2}
Wei Dai.
\newblock Comment on ``singularity summit 2011 workshop report'',
  2012{\natexlab{b}}.
\newblock URL
  \url{https://www.lesswrong.com/posts/acxdcBuZPkvnM9TNM/singularity-summit-2011-workshop-report?commentId=Y4kmaFFFCBoP8HrDS}.

\bibitem[Dai(2012{\natexlab{c}})]{dai2012workshopcomment3}
Wei Dai.
\newblock Comment on ``singularity summit 2011 workshop report'',
  2012{\natexlab{c}}.
\newblock URL
  \url{https://www.lesswrong.com/posts/acxdcBuZPkvnM9TNM/singularity-summit-2011-workshop-report?commentId=5sPXTBRWLmxN6DCf2}.

\bibitem[Dai(2015)]{dai2015amacomment}
Wei Dai.
\newblock Comment on ``{I} am {Nate} {Soares}, {AMA}!'', 2015.
\newblock URL
  \url{https://forum.effectivealtruism.org/posts/cuB3GApHqLFXG36C6/i-am-nate-soares-ama?commentId=4BJn9WkEBrgixL6Bu}.

\bibitem[Dai(2019{\natexlab{a}})]{dai2019purposesdecision}
Wei Dai.
\newblock On the purposes of decision theory research, 2019{\natexlab{a}}.
\newblock URL
  \url{https://www.alignmentforum.org/posts/JSjagTDGdz2y6nNE3/on-the-purposes-of-decision-theory-research}.

\bibitem[Dai(2019{\natexlab{b}})]{dai2019purposesdecisioncomment}
Wei Dai.
\newblock Comment on ``on the purposes of decision theory research'',
  2019{\natexlab{b}}.
\newblock URL
  \url{https://www.lesswrong.com/posts/JSjagTDGdz2y6nNE3/on-the-purposes-of-decision-theory-research/comment/GBZt6scwpAseroYjR}.

\bibitem[Demski(2018)]{demski2018realismcomment}
Abram Demski.
\newblock Comment on ``realism about rationality'', 2018.
\newblock URL
  \url{https://www.lesswrong.com/posts/suxvE2ddnYMPJN9HD/realism-about-rationality/comment/XEbNPAyvjpTcGufLm}.

\bibitem[Demski(2020)]{demski2020realismcomment}
Abram Demski.
\newblock Comment on ``realism about rationality'', 2020.
\newblock URL
  \url{https://www.lesswrong.com/posts/suxvE2ddnYMPJN9HD/realism-about-rationality/comment/3f6fadtcv8FsgwKsz}.

\bibitem[Demski and Garrabrant(2019)]{demski2019embedded}
Abram Demski and Scott Garrabrant.
\newblock Embedded agency.
\newblock \emph{arXiv preprint arXiv:1902.09469}, 2019.
\newblock URL \url{https://arxiv.org/abs/1902.09469}.

\bibitem[Dewar et~al.(1993)Dewar, Builder, Hix, and Levin]{MR-114-A}
James~A. Dewar, Carl~H. Builder, William~M. Hix, and Morlie Levin.
\newblock \emph{Assumption-Based Planning: A Planning Tool for Very Uncertain
  Times}.
\newblock RAND Corporation, Santa Monica, CA, 1993.

\bibitem[Dewey(2017)]{dewey2017hrad}
Daniel Dewey.
\newblock My current thoughts on {MIRI's} ``highly reliable agent design''
  work, 2017.
\newblock URL
  \url{https://forum.effectivealtruism.org/posts/SEL9PW8jozrvLnkb4/my-current-thoughts-on-miri-s-highly-reliable-agent-design}.

\bibitem[Evans et~al.(2021)Evans, Cotton-Barratt, Finnveden, Bales, Balwit,
  Wills, Righetti, and Saunders]{evans2021truthful}
Owain Evans, Owen Cotton-Barratt, Lukas Finnveden, Adam Bales, Avital Balwit,
  Peter Wills, Luca Righetti, and William Saunders.
\newblock Truthful {AI}: Developing and governing {AI} that does not lie, 2021.

\bibitem[Garrabrant et~al.(2016)Garrabrant, Benson-Tilsen, Critch, Soares, and
  Taylor]{garrabrant2016logical}
Scott Garrabrant, Tsvi Benson-Tilsen, Andrew Critch, Nate Soares, and Jessica
  Taylor.
\newblock Logical induction.
\newblock \emph{arXiv preprint arXiv:1609.03543}, 2016.

\bibitem[Krakovna(2017)]{krakovna2017portfolio}
Victoria Krakovna.
\newblock Portfolio approach to {AI} safety research, 2017.
\newblock URL
  \url{https://vkrakovna.wordpress.com/2017/08/16/portfolio-approach-to-ai-safety-research/}.

\bibitem[LaVictoire et~al.(2014)LaVictoire, Fallenstein, Yudkowsky, Barasz,
  Christiano, and Herreshoff]{lavictoire2014program}
Patrick LaVictoire, Benja Fallenstein, Eliezer Yudkowsky, Mihaly Barasz, Paul
  Christiano, and Marcello Herreshoff.
\newblock Program equilibrium in the prisoner's dilemma via {L{\"o}b's}
  theorem.
\newblock In \emph{Workshops at the Twenty-Eighth AAAI Conference on Artificial
  Intelligence}, 2014.

\bibitem[Liskov et~al.(1986)Liskov, Guttag, et~al.]{liskov1986abstraction}
Barbara Liskov, John Guttag, et~al.
\newblock \emph{Abstraction and specification in program development}, volume
  180.
\newblock MIT press Cambridge, 1986.

\bibitem[Ngo(2019)]{ngo2019disentangling}
Richard Ngo.
\newblock Disentangling arguments for the importance of ai safety, 2019.
\newblock URL
  \url{https://www.alignmentforum.org/posts/JbcWQCxKWn3y49bNB/disentangling-arguments-for-the-importance-of-ai-safety}.

\bibitem[Ngo(2020)]{ngo2020plausible}
Richard Ngo.
\newblock Comment on ``plausible cases for {HRAD} work, and locating the crux
  in the `realism about rationality' debate'', 2020.
\newblock URL
  \url{https://www.alignmentforum.org/posts/BGxTpdBGbwCWrGiCL/plausible-cases-for-hrad-work-and-locating-the-crux-in-the?commentId=PfLEwrQfgynqdDH9C}.

\bibitem[Pinker(2018{\natexlab{a}})]{pinker2018enlightment}
Steven Pinker.
\newblock \emph{Enlightenment Now: The Case for Reason, Science, Humanism, and
  Progress}.
\newblock Viking, 2018{\natexlab{a}}.
\newblock ISBN 9780525427575.

\bibitem[Pinker(2018{\natexlab{b}})]{pinker2018robot}
Steven Pinker.
\newblock We’re told to fear robots. {But} why do we think they’ll turn on
  us?, 2018{\natexlab{b}}.
\newblock URL \url{https://www.popsci.com/robot-uprising-enlightenment-now/}.

\bibitem[Rice(2020)]{rice2020}
Issa Rice.
\newblock Plausible cases for {HRAD} work, and locating the crux in the
  ``realism about rationality'' debate, 2020.
\newblock URL
  \url{https://www.alignmentforum.org/posts/BGxTpdBGbwCWrGiCL/plausible-cases-for-hrad-work-and-locating-the-crux-in-the}.

\bibitem[Shah(2020{\natexlab{a}})]{shah2020realismcomment}
Rohin Shah.
\newblock Comment on ``realism about rationality'', 2020{\natexlab{a}}.
\newblock URL
  \url{https://www.lesswrong.com/posts/suxvE2ddnYMPJN9HD/realism-about-rationality/comment/iSubbXvKW7uM6rus6}.

\bibitem[Shah(2020{\natexlab{b}})]{shah2020realismcomment2}
Rohin Shah.
\newblock Comment on ``realism about rationality'', 2020{\natexlab{b}}.
\newblock URL
  \url{https://www.lesswrong.com/posts/suxvE2ddnYMPJN9HD/realism-about-rationality/comment/YMNwHcPNPd4pDK7MR}.

\bibitem[Soares(2017)]{soares2017hradcomment}
Nate Soares.
\newblock Comment on ``my current thoughts on {MIRI's} `highly reliable agent
  design' work'', 2017.
\newblock URL
  \url{https://forum.effectivealtruism.org/posts/SEL9PW8jozrvLnkb4/my-current-thoughts-on-miri-s-highly-reliable-agent-design?commentId=D3PDv7kqJuByt8TRr}.

\bibitem[Soares(2018)]{soares2018update}
Nate Soares.
\newblock Update: Our new research directions.
\newblock Technical report, Machine Intelligence Research Institute, 2018.
\newblock URL
  \url{https://intelligence.org/2018/11/22/2018-update-our-new-research-directions}.

\bibitem[Soares and Fallenstein(2017)]{soares2017agent}
Nate Soares and Benya Fallenstein.
\newblock Agent foundations for aligning machine intelligence with human
  interests: A technical research agenda.
\newblock In V.~Callaghan, J.~Miller, R.~Yampolskiy, and S.~Armstrong, editors,
  \emph{The Technological Singularity: Managing the Journey}. Springer, 2017.
\newblock URL \url{https://intelligence.org/files/TechnicalAgenda.pdf}.

\bibitem[Spolsky(2002)]{spolsky2002}
Joel Spolsky.
\newblock The law of leaky abstractions, 2002.
\newblock URL
  \url{https://www.joelonsoftware.com/2002/11/11/the-law-of-leaky-abstractions/}.

\bibitem[Taylor(2017)]{taylor2017paulmiri}
Jessica Taylor.
\newblock My current take on the {Paul}--{MIRI} disagreement on alignability of
  messy {AI}, 2017.
\newblock URL
  \url{https://www.alignmentforum.org/posts/5bd75cc58225bf06703752c6/my-current-take-on-the-paul-miri-disagreement-on-alignability-of-messy-ai}.

\bibitem[Wentworth(2020)]{wentworth2020plausible}
John Wentworth.
\newblock Comment on ``plausible cases for {HRAD} work, and locating the crux
  in the `realism about rationality' debate'', 2020.
\newblock URL
  \url{https://www.lesswrong.com/posts/BGxTpdBGbwCWrGiCL/plausible-cases-for-hrad-work-and-locating-the-crux-in-the?commentId=Y4uEnMjy9uSXcLswe}.

\bibitem[Wentworth(2021)]{wentworth2021prediction}
John Wentworth.
\newblock Private correspondence re: prediction, December 2021.

\bibitem[Yampolskiy(2016)]{yampolskiy2016leakproofing}
Roman~V Yampolskiy.
\newblock Leakproofing singularity-artificial intelligence confinement problem.
\newblock \emph{Journal of Consciousness Studies JCS}, 2016.

\bibitem[Yampolskiy(2018)]{yampolskiy2018artificial}
Roman~V Yampolskiy.
\newblock \emph{Artificial intelligence safety and security}.
\newblock CRC Press, 2018.

\bibitem[Yampolskiy(2021)]{yampolskiy2021ai}
Roman~V. Yampolskiy.
\newblock {AI} risk skepticism, 2021.
\newblock URL \url{https://arxiv.org/abs/2105.02704}.

\bibitem[Yudkowsky(2006)]{yudkowsky2006martial}
Eliezer Yudkowsky.
\newblock The martial art of rationality, 2006.
\newblock URL
  \url{https://www.lesswrong.com/posts/teaxCFgtmCQ3E9fy8/the-martial-art-of-rationality}.

\bibitem[Yudkowsky(2016)]{yudkowsky2016prereview}
Eliezer Yudkowsky.
\newblock Pre-review comments by {Eliezer} {Yudkowsky}.
\newblock In Nate Soares, editor, \emph{Comments on the Open Philanthropy
  Project's Anonymized Reviews of Three Recent MIRI Papers}. Machine
  Intelligence Research Institute, 2016.
\newblock URL \url{https://intelligence.org/files/OpenPhil2016Supplement.pdf}.

\bibitem[Yudkowsky(2017)]{yudkowsky2017security}
Eliezer Yudkowsky.
\newblock Security mindset and ordinary paranoia, 2017.
\newblock URL
  \url{https://www.lesswrong.com/posts/8gqrbnW758qjHFTrH/security-mindset-and-ordinary-paranoia}.

\bibitem[Yudkowsky(2018)]{yudkowsky2018rocket}
Eliezer Yudkowsky.
\newblock The rocket alignment problem, 2018.
\newblock URL \url{https://intelligence.org/2018/10/03/rocket-alignment/}.

\bibitem[Yudkowsky and Soares(2018)]{yudkowsky2018functional}
Eliezer Yudkowsky and Nate Soares.
\newblock Functional decision theory: A new theory of instrumental rationality,
  2018.
\newblock URL \url{https://arxiv.org/abs/1710.05060}.

\end{thebibliography}

\newpage
\appendix
\section{Selected quotes about deconfusion} \label{deconfusionquotes}
% This section will include all the quotes that are too long to fit in the main body of the paper.
% That way we can say stuff like ``as laid out in quote 7 in Appendix A'' in the main body.

% Issa: In reality, so far, the only long quotes I've wanted to use are from the deconfusion section, so I am framing the appendix in this way. If we find quotes about other things, we can convert this to a more general quotes appendix.
In this section, we give several quotes about deconfusion which provide the basis for our summaries in Section \ref{deconfusionpro}.

\citet{yudkowsky2016prereview}:
\begin{quote}
    Techniques you can actually adapt in a safe AI, come the day, will probably have very simple cores---the sort of core concept that takes up three paragraphs, where any reviewer who didn't spend five years struggling on the problem themselves will think, ``Oh I could have thought of that.'' Someday there may be a book full of clever and difficult things to say about the simple core---contrast the simplicity of the core concept of causal models, versus the complexity of proving all the clever things Judea Pearl had to say about causal models. But the planetary benefit is mainly from posing understandable problems crisply enough so that people can see they are open, and then from the simpler abstract properties of a found solution---complicated aspects will not carry over to real AIs later.
\end{quote}

\citet{bensinger2018comment}:
\begin{quote}
    We're working on decision theory because there's a cluster of confusing issues here (e.g., counterfactuals, updatelessness, coordination) that represent a lot of holes or anomalies in our current best understanding of what high-quality reasoning is and how it works.
\end{quote}

\citet{demski2018realismcomment}:
\begin{quote}
    I don't think there's a true rationality out there in the world, or a true decision theory out there in the world, or even a true notion of intelligence out there in the world. I work on agent foundations because there's \emph{still something I'm confused about} even after that, and furthermore, AI safety work seems fairly hopeless while still so radically confused about the-phenomena-which-we-use-intelligence-and-rationality-and-agency-and-decision-theory-to-describe.
\end{quote}

\citet{soares2017hradcomment}:
\begin{quote}
    The main case for HRAD problems is that we expect them to help in a gestalt way with many different known failure modes (and, plausibly, unknown ones). E.g., `developing a basic understanding of counterfactual reasoning improves our ability to understand the first AGI systems in a general way, and if we understand AGI better it's likelier we can build systems to address deception, edge instantiation, goal instability, and a number of other problems'.
\end{quote}

\section{Clarification of prediction}
\label{clarificationprediction}

In this appendix, we provide a quote from \citet{wentworth2021prediction} that clarifies the meaning of prediction. \citet{wentworth2021prediction}:
\begin{quote}
    I can try to clarify a bit: it's ``predictive'' in the same way that stat mech is predictive, after you already know the underlying laws of quantum mechanics. Logically speaking, stat mech doesn't add anything; we could in-principle calculate everything from quantum mechanics. But practically speaking, we can't predict much of interest from low-level first principles without stat mech. The sort of theory I want would be ``predictive'' in roughly the same sense.

    (That's not necessarily the main intended use-case, but it does convey how ``prediction'' is a tricky term.)
\end{quote}

\end{document}